\documentclass[letterpaper, 10 pt, journal, twoside]{IEEEtran}
\usepackage{cite}

\ifCLASSINFOpdf
\else
\fi
\usepackage{amsmath}
\usepackage{algorithmic}

\usepackage{array}

\usepackage{subfigure}

\usepackage{stfloats}
\usepackage{url}

\usepackage{graphicx}

\usepackage{subfigure}
\usepackage{mathtools}
\usepackage{amssymb}
\usepackage{amsthm}
\usepackage{array}
\usepackage{epstopdf}
\usepackage{xcolor}
\usepackage{multirow}

\usepackage[normalem]{ulem}

\renewcommand{\Re}{\mathbb{R}}

\usepackage{xcolor}
\usepackage{ulem}

\begin{document}
	%
 	\title{Generative Adversarial Networks for Solving Hand-Eye Calibration without Data Correspondence}
	%
	%
	%
	
        \author{Ilkwon Hong$^{1}$ and Junhyoung Ha$^{2}$,~\IEEEmembership{Member,~IEEE}
            \thanks{This work was supported by the National Research Foundation (NRF) of Korea under the grant No. 2022R1C1C1005483 funded by the Korea government (MSIT). {\it (Corresponding author: Junhyoung Ha.)}}
            \thanks{{$^1$}Ilkwon Hong is with the RoboticsLAB, Hyundai Motor Group, Gyeonggi 16082, South Korea (e-mail: {\tt\small ikhong@hyundai.com})}
            \thanks{{$^2$}Junhyoung Ha is with the Artificial Intelligence and Robotics Institute, Korea Institute of Science and Technology, Seoul 02792, South Korea (e-mail: {\tt\small hjhdog1@gmail.com}).}
            \thanks{This work has been submitted to the IEEE for possible publication. Copyright may be transferred without notice, after which this version may no longer be accessible.}
        }
	\maketitle
	
 

\begin{abstract}
In this study, we rediscovered the framework of generative adversarial networks (GANs) as a solver for calibration problems without data correspondence. When data correspondence is not present or loosely established, the calibration problem becomes a parameter estimation problem that aligns the two data distributions. This procedure is conceptually identical to the underlying principle of GAN training in which networks are trained to match the generative distribution to the real data distribution. As a primary application, this idea is applied to the hand-eye calibration problem, demonstrating the proposed method's applicability and benefits in complicated calibration problems.
\end{abstract}


	%


\section{Introduction}
Calibration is an essential process in many engineering applications where unknown system parameters must be estimated. Integration of multiple measurement systems is one example, in which unknown parameters are introduced to define the relative configurations of the individual systems.  A common scenario in robotics is the {\it `hand-eye calibration'}, which involves estimating an unknown transformation between a robot hand and a camera.

The hand-eye calibration problem is generally defined by two distinct loop-closure equations, the $AX=XB$ and $AX=YB$, which are comprised of elements of $SE(3)$, the Euclidean group of rigid body motions. In the $AX=XB$ formulation, $X \in SE(3)$ is the only unknown transformation, whereas, in the $AX=YB$ formulation, there are two unknown transformations $X \in SE(3)$ and $Y \in SE(3)$. The unknown transformations can be estimated using the measurement sets $\{A_i\}$ and $\{B_j\}$, where $A_i$ and $B_j$ are also elements of $SE(3)$.

Most studies assume that the exact correspondence between $A_i$ and $B_j$ exits and is known. The dataset in this scenario was expressed as a set of measurement pairs $\{(A_i, B_i)\}$.  Extensive studies have been conducted on hand-eye calibration under this assumption, including the derivation of algebraic and linear least-squares solutions~\cite{park1994robot,dornaika1998simultaneous,ha2015stochastic,condurache2019novel,shah2013solving,ernst2012non} and nonlinear optimization approaches~\cite{dornaika1998simultaneous,li2010simultaneous,tabb2017solving,ernst2012non}. Recently, deep learning-based approaches have been introduced in \cite{hua2021hand,bahadir2022deep}.

Another problem formulation was suggested under the assumption that the measurement correspondence is known, whereas unknown offsets exist in the acquisition times between $A_i$ and $B_i$~\cite{furrer2018evaluation,pachtrachai2018chess}. These studies seek to address more practical situations in which measurements are collected from distinct devices with asynchronous clocks, especially when the system must be re-calibrated online while in operation. This problem is broader than the first since its method is applicable to both, whereas the reverse is not true.

This suggests a hierarchical perspective toward problem formulation regarding associated data types. Further generalizing, we can consider a problem formulation independent of data correspondence~\cite{ackerman2013probabilistic,ackerman2014information,li2015simultaneous,ma2016new}. In this case, the exact correspondence between $\{A_i\}$ and $\{B_j\}$ may not exist because of the different sampling frequencies. Furthermore, the measurements can be completely scrambled. As matching the data distributions between the left- and right-side transformations is a natural approach in this situation, these studies modeled probability density functions (pdfs) in the space of $SE(3)$ as the {\it `highly focused pdfs,'} and found $X$ or $(X, Y)$ that matches the first two moments of the pdfs, i.e., the mean and covariance.

Instead of relying on lower-order moments, a method that aligns the entire data distributions can be more accurate. It is also beneficial if no explicit distribution modeling is required. We found clues from recent advances in deep learning technology that have attracted enormous attention over the last decade. Generative Adversarial Networks (GANs) are among the most representative techniques used in recent advances in deep learning. It incorporates two neural networks, a  generator and a discriminator, which compete during training. Although there have been many creative variations of GAN, we focused on the very first invention~~\cite{NIPS2014_5ca3e9b1}, the vanilla GAN, where the training phase was formulated as a pdf matching between the generative and real data distributions.

In this study, we rediscovered the GAN framework as a solver for calibration problems without data correspondence. As a primary application, we demonstrate how hand-eye calibration without data correspondence can be solved by exploiting the GAN training principle. Our study focuses on the $AX=XB$ formulation because the $AX=YB$ problem can be reduced to the $AX=XB$ problem by merging every measurement pair. We cast the GAN training to a hand-eye calibration solver by including $X$ as the generator's network parameter. In contrast to the typical use of GAN, the trained generator is not to be used as a generative model. Instead, only the resulting network parameter is taken as the calibrated transformation.

Inheriting the established benefits of the GAN, the advantages of our approach over existing methods are as follows:
\begin{itemize}
    \item No pdf modeling is required. This is particularly beneficial when handling variables in nonlinear manifolds, such as the space of $SE(3)$;
    \item It utilizes the entire data distributions, without relying on finite moments;
    \item It is easily adaptable to other parameter estimation problems without data correspondence.
\end{itemize}
Regarding the third point, we present an algebraic condition under which the proposed method is extendable and also provide an example of polynomial parameter estimation.

The remainder of this paper is organized as follows. The next section discusses prior research on related topics. Section~\ref{sec:problem} defines the problem, followed by a description of the GAN-based calibration method in Section~\ref{sec:method}. In Section~\ref{sec:sim} and Section~\ref{sec:hardware}, the proposed method is validated through a set of numerical and hardware experiments, respectively. Subsequently, Section~\ref{sec:general_param} demonstrates how the proposed method can be extended to general parameter estimation problems. Finally, the conclusions are presented in Section~\ref{sec:conclusions}.

\section{Related Works}
\label{sec:related_works}

Related topics include hand-eye calibration, generative adversarial networks, and divergence measures for pdfs. 

\subsection{Hand-Eye Calibration}

\subsubsection{Under Exact Correspondence}

Given the exact correspondence, the measurement data are expressed as $\{(A_i, B_i)\}$. When noiseless measurements were available, the algebraic solution to the $AX=XB$ formulation was derived by \cite{park1994robot} for two independent pairs of $(A_i, B_i)$. For the $AX=YB$ formulation, the algebraic solutions were presented in \cite{dornaika1998simultaneous,ha2015stochastic,condurache2019novel} for three independent measurement pairs.

When measurement noise exists, a typical approach is to formulate a least-squares minimization:
\begin{equation}
    \min_{X \in SE(3)} \sum_i d(A_i X, X B_i)^2
\end{equation}
or
\begin{equation}
    \min_{X, Y \in SE(3)} \sum_i d(A_i X, Y B_i)^2,
\end{equation}
where $d(\cdot, \cdot)$ is the distance function of $SE(3)$. Various representations of homogeneous transformations have been used to solve the minimization. In \cite{park1994robot}, the Lie theory on the group of rotation matrices $SO(3)$ was accommodated, where the associated Lie algebra $so(3)$ was used to represent the rotations. Unit quaternions were used for rotations in \cite{dornaika1998simultaneous} and \cite{zhuang1994simultaneous}, and dual-quaternion representations were incorporated for homogeneous transformations in \cite{li2010simultaneous}.

In most previous studies, a linear least-squares solution was computed and subsequently refined through nonlinear iterative optimization~\cite{dornaika1998simultaneous,li2010simultaneous,tabb2017solving,ernst2012non}. To find a linear solution, the method in \cite{park1994robot} used the elements of $so(3)$ as $3$D vectors in a Singular Value Decomposition (SVD)-based approach. Other linear solutions have been derived based on the Kronecker product~\cite{shah2013solving} or using a relaxed orthogonality constraint~\cite{ernst2012non}. For nonlinear iterative optimization, an algorithm using an analytic gradient was presented in \cite{ha2015stochastic}, and various iterative methods were proposed and compared in \cite{tabb2017solving}. More recently, a probabilistic interpretation of the $AX=YB$ problem was presented in \cite{ha2022probabilistic} along with an iterative maximum likelihood estimation.
As extensions of the classic hand-eye calibration problem, more advanced solvers have been introduced for scenarios involving multi-camera systems~\cite{wang2022accurate,evangelista2023graph} and robots with uncertainties~\cite{ulrich2021generic}.

\subsubsection{Under Loose Correspondence}

Temporal alignment is additionally required when data streams are asynchronous. In \cite{furrer2018evaluation}, an approach to spatiotemporal estimation for hand-eye calibration was proposed incorporating interpolation-based time alignment and outlier-robust hand-eye calibration. Similarly, a method for data synchronization using cross-correlation and resampling was proposed in \cite{pachtrachai2018chess}.

In the context of general sensor calibration, in \cite{taylor2016motion}, the existing hand-eye calibration is extended to the estimation of multimodal sensor extrinsics and timing offsets using sensor motion information. In \cite{rehder2016general}, an estimation of the constant temporal offsets of the sensors, along with the relative sensor transformation, was proposed in the sensor fusion context. An Iterative Closest Point-based algorithm in three-dimensional orientation space was proposed by \cite{kelly2014general} for spatio-temporal sensor calibration.

\subsubsection{Without Correspondence}

The first studies on hand-eye calibration when the measurement correspondence is unknown were presented in \cite{ackerman2013probabilistic,ackerman2014information}, where the $AX=XB$ problem was considered. The data distributions in $SE(3)$ were modeled as the highly focused distributions, of which the mean and covariance were algebraically computed  through a convolution-based derivation. Subsequently, the unknown transformation $X$ was computed by matching the mean and covariance between $\{A_i X\}$ and $\{X B_j\}$. This approach was further modified by \cite{ma2016new} by exploiting the first- and second-order approximations of the mean in $SE(3)$.

In \cite{li2015simultaneous}, the $AX=YB$ problem was considered by using the same framework. Accordingly, the same derivations of the mean and covariance were used for $\{A_i X\}$ and $\{Y B_j\}$. By matching the mean and covariance, eight candidates of $(X,Y)$ were identified, followed by a process to select the correct pair.

\subsection{Generative Adversarial Networks (GANs)}

The concept of GANs was first introduced in \cite{NIPS2014_5ca3e9b1} to estimate generative models through an adversarial process. The GAN framework consists of two neural networks, a generator and a discriminator, with the aim of training the generator network to accurately capture the given data distribution. During the training process, the discriminator compares the generative distribution to the data distribution and provides guidance to the generator to imitate the data distribution.

Over the last decade, GANs have been studied extensively, mostly for image synthesis and generation~\cite{choi2018stargan,brock2018large,karras2019style}. Recently, it has been considered old-fashioned and has rapidly lost its ground in the image generation field to newer technologies, particularly diffusion-based methods~\cite{dhariwal2021diffusion,rombach2022high}.

\subsection{Kullback–Leibler Divergence}

A classical approach to aligning two pdfs is to minimize the divergence measure. The Kullback–Leibler (KL) divergence is a popular divergence measure between two pdfs:
\begin{equation}
D(p(x)\|q(x))=\int_{\Re^d} p(x) \log \frac{p(x)}{q(x)} dx,
\end{equation}
where $p(x)$ and $q(x)$ denote the pdfs defined in $\Re^d$. An extension of the KL divergence to the space of $SE(3)$ was studied in \cite{chirikjian2010information,chirikjian2011stochastic,chirikjian2010informationtheoretic}. 

When using the KL divergence, analytic pdf models are preferable for easier integration and flexible algebraic manipulation. In \cite{ackerman2014information}, for example, the KL divergence in $SE(3)$ was applied to hand-eye calibration using an algebraic model of the Gaussian distribution in $SE(3)$. 

When only independent and identically distributed (i.i.d.) samples from $p(x)$ and $q(x)$ are provided, the exact KL divergence is not available, while an estimator can be used instead. A popular estimator of the KL divergence in the vector space was presented in \cite{perez2008kullback}. To apply this to hand-eye calibration, it should be extended to the space of $SE(3)$. This extension is not straightforward because the estimator involves defining a uniform distribution and calculating the volume of unit-ball, which is not trivial in $SE(3)$.

\section{Hand-Eye Calibration without Data Correspondence}
\label{sec:problem}

Let us consider $\rho(B)$ as the underlying pdf of $\{ B_j \}$. A transformation $B$ can be sampled from $\rho(B)$ as
\begin{equation}
B \sim \rho(B). \label{eqn:p(B)}
\end{equation}
By rewriting the loop-closure equation as $B = X^{-1} A X$, a transformation $A$ can also be sampled as
\begin{equation}
X^{-1} A X \sim \rho(B). \label{eqn:p(A)}
\end{equation}

In our problem formulation, we assume that the given datasets $\{A_i\}$ and $\{B_j\}$ are i.i.d. samples acquired in this manner. In this situation, the set sizes can differ from each other, and no element-wise correspondence is assumed; not only are the element-wise correspondences unknown, but they also might not exist. 
Because $B$ and $X^{-1} A X$ are both drawn from the same distribution $\rho(B)$, the calibration process is reduced to finding a transformation $X$ that aligns the two data distributions $\{ B_j \}$ and $\{ X^{-1} A_i X \}$.

\section{Generative Adversarial Networks for Hand-Eye Calibration without Data Correspondence}
\label{sec:method}

The vanilla GAN outlined in \cite{NIPS2014_5ca3e9b1} involves training a generator network to produce fake samples that resemble real data.
The training of the generator is guided by an additional neural network called a discriminator, which is trained to distinguish between real and fake samples.

The two networks are trained in an adversarial manner. At each training iteration, the generator creates random fake samples, and the discriminator uses them to improve its ability to distinguish real from fake samples. Afterward, the generator is trained to generate fake samples that can fool the discriminator more effectively. Repeating this process, the generative distribution converges to the real data distribution.

In principle, this process can be outlined as aligning two data distributions, which is precisely what is required for our purposes. In this study, we utilize the GAN training principle for hand-eye calibration without requiring data correspondence information.

\subsection{Method Outline}

\begin{figure}[t]
  \center
    \includegraphics[width=0.9\columnwidth]{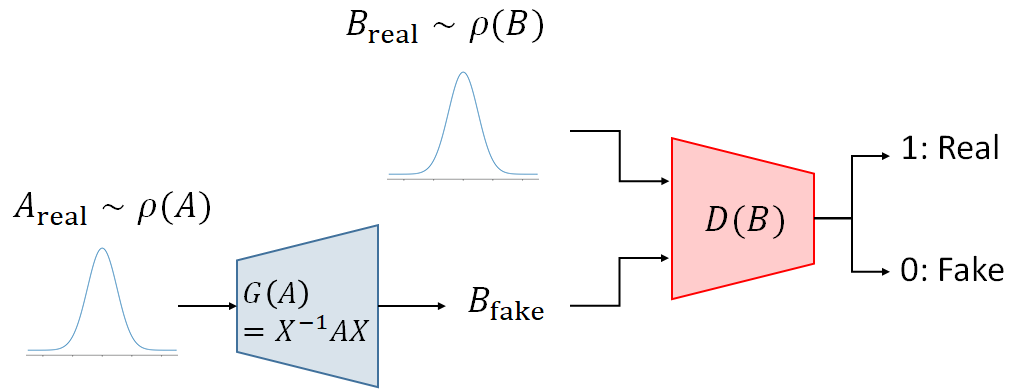}
\caption{Proposed GAN-based method for hand-eye calibration without data correspondence.}
\label{fig:method}
\end{figure}

Our GAN-based calibration method is illustrated in Figure ~\ref{fig:method}, where fake $B$'s are created by the generator $G$, and the real $B$'s are sampled from $\{ B_j \}$. The discriminator $D$ is trained to differentiate between real and fake $B$'s.

In the vanilla GAN, $G$ and $D$ are ideally explored in the space of arbitrary functions, and $G$ takes a noise vector as input. In contrast, the primary differences in our approach are as follows.
\begin{itemize}
    \item The generator $G$ is modeled as
    \begin{equation}
        G(A) = X^{-1} A X,
    \end{equation}
    where $X \in SE(3)$ is the network parameter.
    \item The generator's input is a transformation $A$ randomly sampled from $\{ A_i \}$. By referring to the input and output as $A_\text{real}$ and $B_\text{fake}$, respectively, the generation is expressed as
    \begin{equation}
        B_\text{fake} = G(A_\text{real}).
    \end{equation}
    \item The vanilla GAN framework trains $G$ for later use as a generative model, whereas our approach purely aims at the training process, from which the resulting $X$ is obtained as the calibrated transformation.
\end{itemize}

\subsection{Mode Collapse-Free Characteristic}

One of the most common problems in GAN training is ``mode collapse,'' which refers to the phenomenon that the generator is trained to capture only a part of data distribution. The risk of mode collapse exists when the generator is explored in the space of arbitrary functions.

In our case, our generator is not an arbitrary function but a function of the form  $G(A)=X^{-1} A X$, which theoretically never experiences mode collapse: the generator only left- and right-translates the input distribution $\{A_i\}$, making it impossible for mode collapse to occur. This can also be understood mathematically by $G(A_i) \neq G(A_j)$ if and only if $A_i \neq A_j$.

\subsection{Implementation Details}

Training GAN is known to be challenging and requires considerable effort. In our study, we encountered several practical issues, which we will discuss here along with their remedies.

\subsubsection{Orthogonality-Preserving Updates}

The generator's parameter $X$ is an element of $SE(3)$, which consists of the rotation $R_X \in SO(3)$ and the position $p_X \in \Re^3$. Notably, the rotation $R_X$ is subject to the orthogonality constraint $R^T R = I$. We rely on PyTorch's optimizer to update the position $p_X$, while the rotation $R_X$ must be updated in a way that preserves orthogonality.

A naive approach is to use a three-dimensional vector representation of $SO(3)$ such as Euler angles or the exponential coordinates~\cite{lynch2017modern}. The drawback of this case is that these spaces are not isometric to $SO(3)$. This implies that constant velocities in these spaces at different locations in different directions are not constant in the space of $SO(3)$. Consequently, the effective learning rate can vary unintentionally during the training.

This is a typical issue in iterative algorithms for $SO(3)$. To address this issue, a coordinate-invariant approach was applied using the exponential map as local coordinates for each iterative update~\cite{ha2015stochastic,ha2022probabilistic,park2000geometric}. We use the same approach in which the update is expressed as
\begin{equation}
R_X \leftarrow R_X e^{[w]}. \label{eqn:update}
\end{equation}
Here, $R_X \in SO(3)$ is the rotation of $X$, $w \in \Re^3$ is the update vector, and $[\cdot]$ denotes an element of $so(3)$, i.e., the Lie algebra associated with $SO(3)$, or equivalently, the skew-symmetric matrix representation~\cite{lynch2017modern}.

\subsubsection{Reducing Solution Stochasticity}

The stochastic nature of GAN training, which has been attributed to the mini-batch training~\cite{yazici2020empirical} and the nature of two-player games~\cite{mescheder2017numerics,gidel2018variational}, can affect the solution convergence of our method. Specifically, we observed that the parameter $X$ orbits around the optimal solution during training; therefore, the resulting solution of $X$ depends on when the training is terminated. To address this issue, we adopted the simplest approach of averaging $X$ over multiple training iterations, resulting in reduced randomness.

\subsubsection{Attaining Global Solution}
\label{subsubsec:global}

Another issue we encountered during implementation was local convergence. A popular approach for obtaining a global solution is stochastic global optimization~\cite{rinnooy1987stochastic1,rinnooy1987stochastic2}. The simplest method for stochastic global optimization is multi-start optimization, in which the best solution is obtained after multiple local optimizations with different random initial guesses.

In our case, the best generator must be selected after multiple training executions. When comparing between generators, the quality of each generator can be measured by the similarity between the distributions of the fake and real $B$'s. In the vanilla GAN framework, the discriminator $D$ estimates the probability of a given sample to be real~\cite{NIPS2014_5ca3e9b1}, and a unique solution of $G$ and $D$ exists as $G$ recovering the data distribution and $D$ being equal to $\frac{1}{2}$. If $G$ does not accurately capture the data distribution, the output of $D$ diverges from $\frac{1}{2}$ as training proceeds. Using this property, we define the quality measure of $G$ as 
\begin{equation}
Q = 1 - \mathbb{E} \left[ 2 \left(D(G(A)) - \frac{1}{2} \right)^2 \right] - \mathbb{E} \left[ 2 \left( D(B)  - \frac{1}{2} \right)^2 \right],
\label{eqn:quality}
\end{equation}
which is $0$ in the worst case and $1$ in the best case. Although $Q$ is not reliable in the early stages of training because $D$ is generally initialized as $D \approx 0.5$ (leading to $Q \approx 1$), the final $Q$ after sufficient training provides a good evaluation of the estimated $X$. For computational efficiency, we terminate the multi-start optimization once we achieve a $Q$ surpassing a chosen quality threshold.

\subsubsection{Data Normalization}
Training neural networks generally requires prior data normalization.
Notably, our $SE(3)$ data contained rotations and positions. The elements of rotation are always between $-1$ and $1$ by definition, whereas the position scale varies with the dataset. Therefore, we propose an SVD-based method to normalize the position vectors in datasets $\{ A_i \}$ and $\{ B_j \}$ according to the maximum expected scale of the position vector of $X$.

By taking the average of the equation $AX=XB$, we obtain
\begin{equation}
\Bar{A}X = X\Bar{B} \label{eqn:mean_axxb}
\end{equation}
where $\Bar{A}$ and $\Bar{B}$ are the arithmetic means of $A$ and $B$. Thus, they are not necessarily elements of $SE(3)$. Given the datasets $\{A_i\}$ and $\{B_j\}$, $\Bar{A}$ and $\Bar{B}$ are calculated as
\begin{equation}
\begin{split}
\Bar{A} = \left[ \begin{array}{cc} 
\Bar{R}_A & \Bar{p}_A \\ 0_{1 \times 3} & 1
\end{array} \right] = {1\over n}\sum_{i=1}^n {A_i}
\\
\Bar{B} = \left[ \begin{array}{cc} 
\Bar{R}_B & \Bar{p}_B \\ 0_{1 \times 3} & 1
\end{array} \right] = {1\over m}\sum_{j=1}^m {B_j}
\end{split}
\label{eqn:barA_barB}
\end{equation}
where $n$ and $m$ are the numbers of elements in $\{A_i\}$ and $\{B_j\}$, respectively, and $\Bar{R}_{(\cdot)}$ and $\Bar{p}_{(\cdot)}$ are the means of the rotation and position of the subscripted variable, respectively.

After an algebraic manipulation with (\ref{eqn:mean_axxb}) and (\ref{eqn:barA_barB}), the position vector of $X$ is given by
\begin{equation}
p_X = \left( \bar{R}_A - I_{3 \times 3}\right)^{-1}\left(R_X \Bar{p}_B - \Bar{p}_A \right).
\label{eqn:pX}
\end{equation}
Subsequently, the following inequality holds for $\| p_X \|$:
\begin{equation}
\begin{split}
\| p_X \| &=
\| \left( \bar{R}_A - I_{3 \times 3}\right)^{-1}\left(R_X \Bar{p}_B - \Bar{p}_A \right) \| \\
&\le \sigma_\text{max} \| R_X \Bar{p}_B - \Bar{p}_A \| \\
&\le \sigma_\text{max} \left( \| \Bar{p}_B \| + \| \Bar{p}_A \| \right)
\label{eqn:px_upperlimit}
\end{split}
\end{equation}
where $\sigma_\text{max}$ is the maximum singular value of $\left( \bar{R}_A - I_{3 \times 3}\right)^{-1}$. Finally, the upper limit of $\| p_X \|$ is given as $\sigma_\text{max} \left( \| \Bar{p}_B \| + \| \Bar{p}_A \| \right)$, which is used in our method as a normalization factor for the position vectors.
\subsubsection{Discriminator Architecture}
Our discriminator model takes the sixteen elements in $B$ and propagates them through linear, batch-norm., and Leaky ReLU layers, producing a single real number ranging between $0$ and $1$. The detailed layer structures are as follows:
\begin{itemize}
    \item The model incorporates seven linear layers, as detailed in Table~\ref{tab:D};
    \item Batch normalization is applied only to Linear Layer $3$;
    \item The activation function used is Leaky ReLU with a slope of $0.1$;
    \item Dropout with a probability of $0.5$ is applied to all linear layers except the last layer;
    \item The final output is processed through a sigmoid function.
\end{itemize}


\begin{table}
\caption{Linear layers in discriminator network}
\label{tab:D}
\begin{center}
\begin{tabular}{r|c|c|c|c|c|c|c}
\cline{2-8}
\multicolumn{1}{l}{} & \multicolumn{7}{c}{\bf Linear Layers} \\
\cline{2-8}
\multicolumn{1}{l}{} & {\bf 1} & {\bf 2} & {\bf 3} & {\bf 4} & {\bf 5} & {\bf 6} & {\bf 7} \\
\hline
\hline
{\bf Input Channel \#} & {16} & {64}  & {128} & {128} & {256} & {128} & {64} \\
\hline
{\bf output Channel \#} & {64} & {128}  & {128} & {256} & {128} & {64} & {1} \\
\hline
\end{tabular}
\end{center}
\end{table}


\section{Numerical Simulations}
\label{sec:sim}

We performed several numerical simulations to compare our method with the existing analytical methods in \cite{ackerman2013probabilistic} and \cite{ma2016new}. We refer to them as `batch method' and `new batch method (1st order),’ respectively.

\subsection{Synthetic Data Generation}
\label{subsec:data}

Synthetic noiseless datasets $\{ A_i \}$ and $\{ B_j \}$ were generated as follows.

First, we randomly generate $X_\text{true}$, the ground-truth transformation of $X$, by
\begin{equation}
\begin{split}
&X_\text{true} = \left[ \begin{array}{cc}
R & d \hat{p} \\ 0_{1 \times 3} & 1
\end{array} \right], \\
&R \sim U_{SO(3)}(R), \ 
\hat{p} \sim U_{S^2}(\hat{p}),
\end{split}
\end{equation}
where $U_{SO(3)}(R)$ and $U_{S^2}(\hat{p})$ are the uniform distributions on $SO(3)$ and $S^2$ (the $2$-sphere), respectively, and $d \in \Re$ is the position scale. To simulate realistic situations, we used $d = 125.31$ mm, which is the position scale of the hand-eye transformation observed in our later hardware experiment. We refer to \cite{ha2022probabilistic} for a uniform sampling method on $SO(3)$, whereas sampling on $S^2$ is easily performed by normalizing an isotropic Gaussian random vector.

Subsequently, $B$'s are randomly generated, and $A$'s are calculated accordingly by $A = X_\text{true} B X_\text{true}^{-1}$. A more precise procedure is as follows:
\begin{equation}
\begin{split}
&B_k = B_0 \left[ \begin{array}{cc}
e^{[w_k]} & d p_k  \\ 0_{1 \times 3} & 1
\end{array} \right], \\
&w_k \sim \mathcal{N}(0;\Sigma_w), \ p_k \sim \mathcal{N}(0;\Sigma_p)
\end{split}
\label{eqn:simulation data sampling}
\end{equation}
and
\begin{equation}
A_k = X_\text{true} B_k X_\text{true}^{-1} \ \text{ for } \ k = 1, \ldots, n+m,
\end{equation}
where $\Sigma_w$ and $\Sigma_p$ are random diagonal matrices, whose diagonal elements are sampled from $[0,1]$. Again, the position vector of $B_k$ is scaled by $d$ to reflect the physical scale. The baseline transformation $B_0 \in SE(3)$ is created by sampling a uniform random rotation, as we did with $X_\text{true}$, and a position vector as an isotropic Gaussian random vector scaled by $10 d$.

Once $n+m$ pairs  $\{ (A_k, B_k) \}$ are generated, they are separated into two sets with set sizes $n$ and $m$. Subsequently, only $A$'s are taken from the first set, and only $B$'s are taken from the second set, resulting in noiseless datasets $\{ A_i \}$ and $\{ B_j \}$. This process ensures that no element-wise correspondence exists between the sets.

When noisy datasets are required, we add right-translated noise to each transformation in $\{ A_i \}$ and in $\{ B_j \}$, as we generated $B_k$ by right-translating $B_0$ with a random transformation, as given in (\ref{eqn:simulation data sampling}).

\subsection{Simulation Results}

We conducted $100$ noiseless simulations and $100$ noisy simulations to compare the performance of our method and the existing methods. In each simulation, the datasets were generated independently, as described in Section~\ref{subsec:data}, with randomly selected $\Sigma_w$ and $\Sigma_p$. The set sizes of $\{ A_i \}$ and $\{ B_i \}$ were chosen as $n = 6000$ and $m = 4000$, respectively, in all the simulations.

To evaluate the estimation accuracy, two errors were considered: the rotation error, defined as the angular displacement with respect to $X_\text{true}$, and the translation error, defined as the positional distance from $X_\text{true}$.

\begin{figure}[t]
  \subfigure[]{
    \includegraphics[width=0.46\columnwidth]{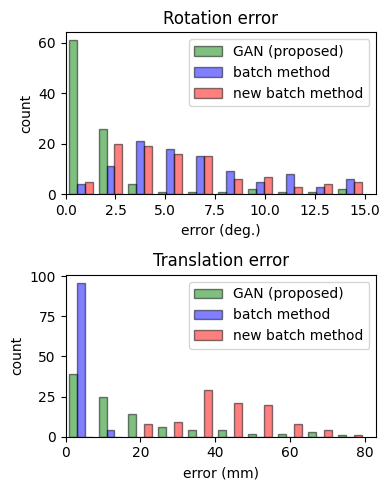}
  }
  \subfigure[]{
    \includegraphics[width=0.46\columnwidth]{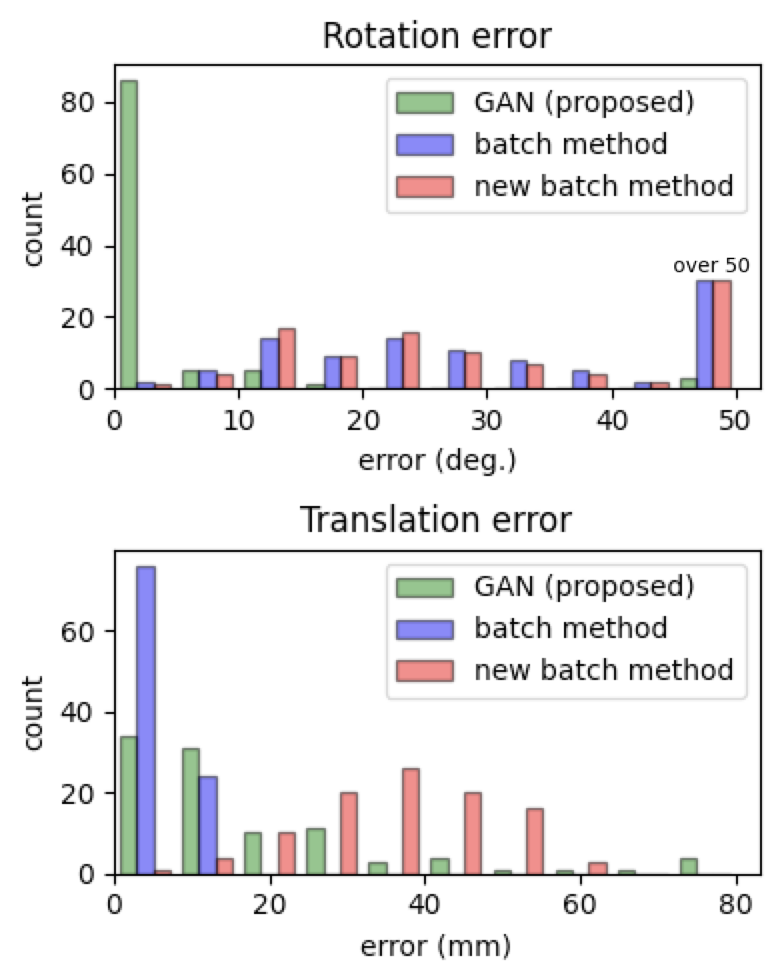}
  }
\caption{Error comparison between proposed and existing methods in numerical simulations (a) using noiseless data and (b) using noise data.}
\label{fig:simulation result split}
\end{figure}

The simulation errors were compared between the methods, as shown in the histograms in Fig.~\ref{fig:simulation result split}. As observed, our method outperformed the other methods in terms of rotation error, while the batch method was the most effective in minimizing translation error. This can be attributed to the solution stochasticity of our method, which was observed more prominent in translation than in rotation in our implementation. Given that our implementation was not fine-tuned and primarily focused on demonstrating the feasibility of GAN-based calibration, we believe that this issue can be addressed by further exploring optimal learning rates and hyperparameters.

Notably, our method exhibited less degradation compared to the other methods in noisy simulations. This would be because the compared methods were derived by strictly equating the first two moments between two data distributions; thus, they are sensitive to factors that may break the equality. On the other hand, our method aligns the distributions themselves, making it more capable of regressing noise.

For each simulation, our method took $11.3$ seconds on average on the Apple silicon M2 using PyTorch with CPU computation.

\section{Hardware Experiment}
\label{sec:hardware}

To demonstrate a real-world comparison between our method and the existing methods, we performed hand-eye calibration using a robot arm (Franka Emika Panda) and an RGB-camera attached to the end-effector of the robot through a $3$D-printed fixture. The camera transformations were acquired by capturing a checkerboard using a built-in Perspective-n-Point (PnP) solver in OpenCV library, and the robot transformations were acquired by robot kinematics, as shown in Fig.~\ref{fig:hardware}.

\subsection{Experimental Data Acquisition}

\begin{figure}[t]
  \center
    \includegraphics[width=0.6\columnwidth]{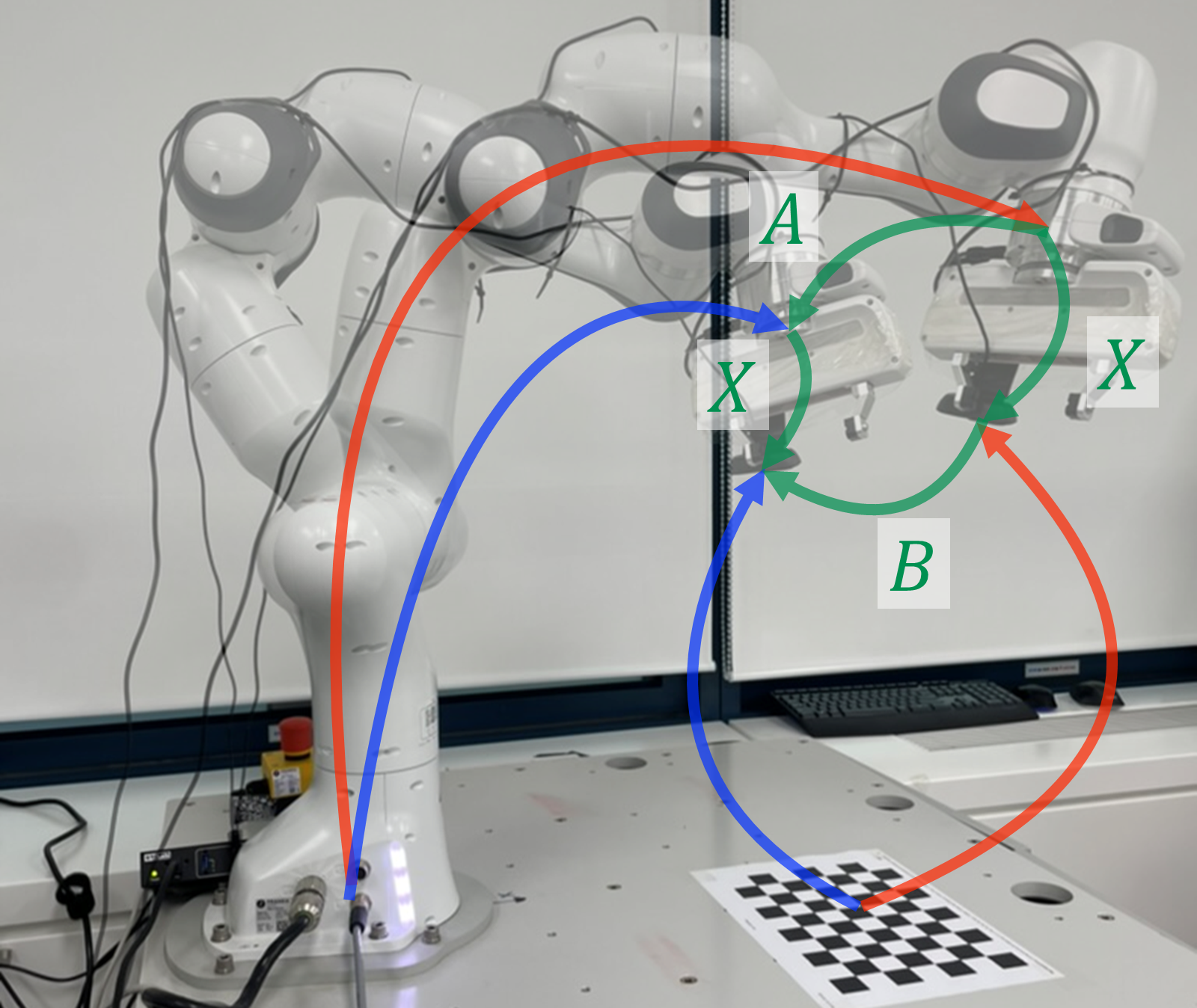}
  
\caption{
Two system configurations constitute the $AX=XB$ loop in our experimental setup. An RGB-camera was attached to the end-effector of a Franka Emika Panda manipulator and captured a fixed checkerboard. }
\label{fig:hardware}
\end{figure}

In this experiment, the robot moved continuously for approximately $3$ min, and the robot and camera transformations were recorded independently at different sampling frequencies. To constitute the $AX=XB$ loop, two distinct configurations must be merged, as indicated by the green loop in Fig.~\ref{fig:hardware}. Because there was no correspondence between the robot and camera transformations, we merged every two robot and two camera transformations to obtain $\{ A_i \}$ and $\{ B_j \}$, respectively.

We conducted three independent experiments, denoted as Exp.~1, Exp.~2, and Exp.~3, with the camera mounted at different positions on the robot's end-effector. In each experiment, we acquired $a$ robot and $b$ camera transformations, recored as $(a, b) = (626, 696)$, $(626, 479),$ and $(669, 540)$, respectively. This resulted in the set sizes of $\{ A_i \}$ and $\{ B_j \}$ being $(n,m) = (391,250, \ 483,720)$, $(391,250, \ 228,962)$, and $(446,892, \ 291,060)$ for the three experiments, respectively. These numbers represent all possible pairwise combinations of the measurement transformations.

For each experiment, the ground truth of $X$ was obtained by collecting a dataset with known correspondences, $\{ (A_i, B_i) \}$, and applying a conventional hand-eye calibration method under correspondence~\cite{ha2015stochastic}. 

\subsection{Results}

\begin{figure}[t]
  \center
    \includegraphics[width=1\columnwidth]{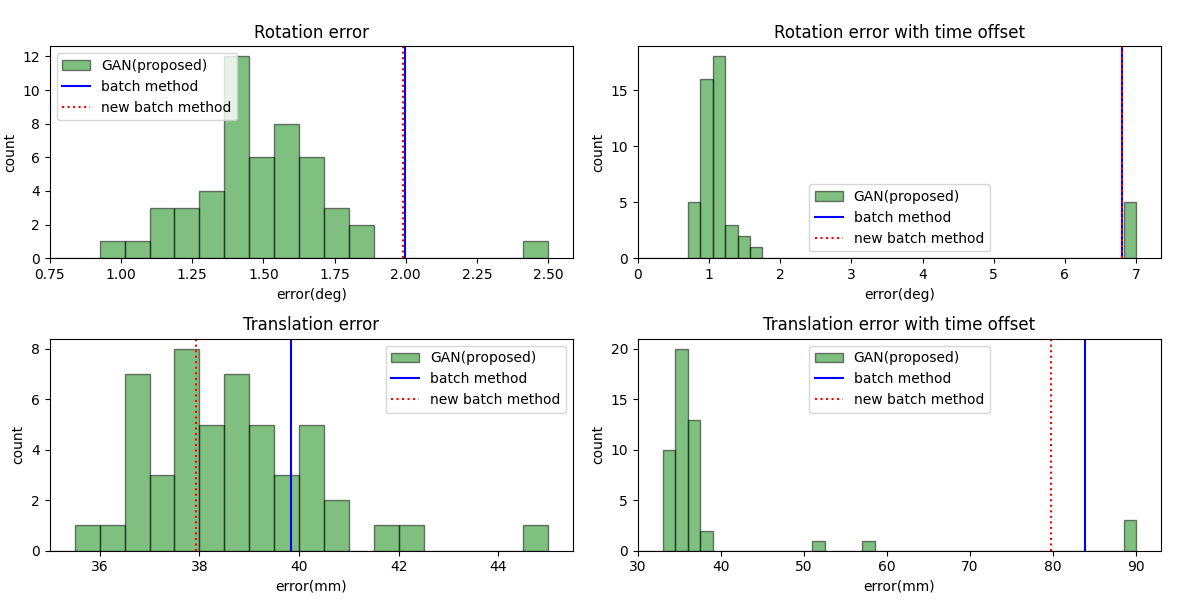}
\caption{Error comparison for Exp.~1 between proposed and existing methods using experimental data without time offsets (left column) and with simulated time offsets (right column).}
\label{fig:exp1}
\end{figure}

\begin{figure}[t]
  \center
    \includegraphics[width=1\columnwidth]{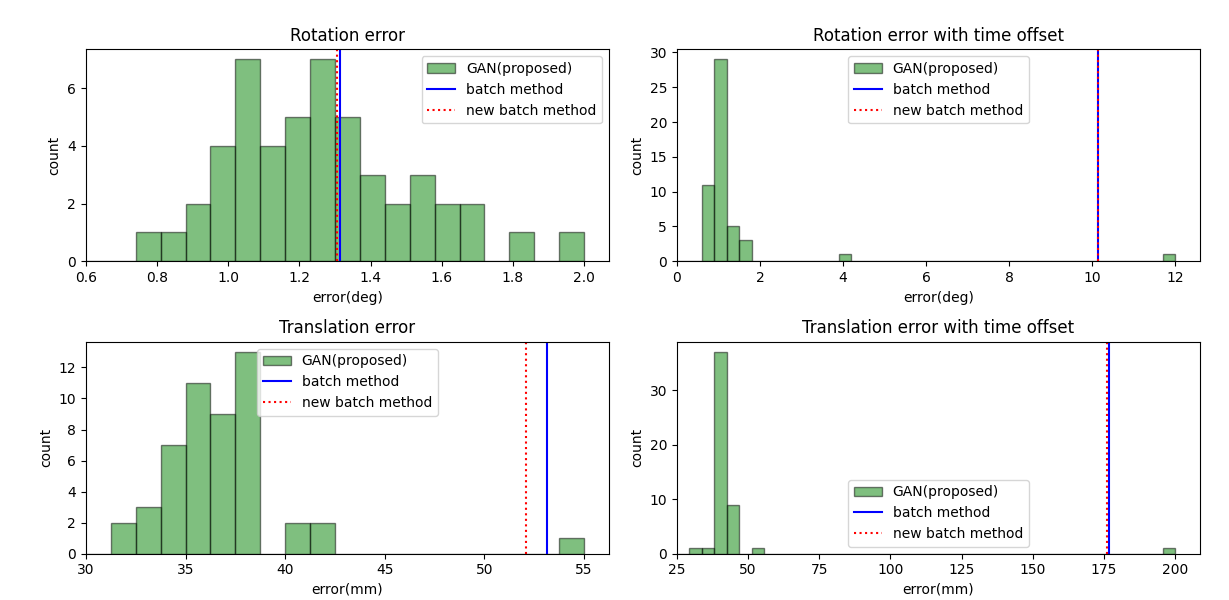}
\caption{Error comparison for Exp.~2 between proposed and existing methods using experimental data without time offsets (left column) and with simulated time offsets (right column).}
\label{fig:exp2}
\end{figure}

\begin{figure}[t]
  \center
    \includegraphics[width=1\columnwidth]{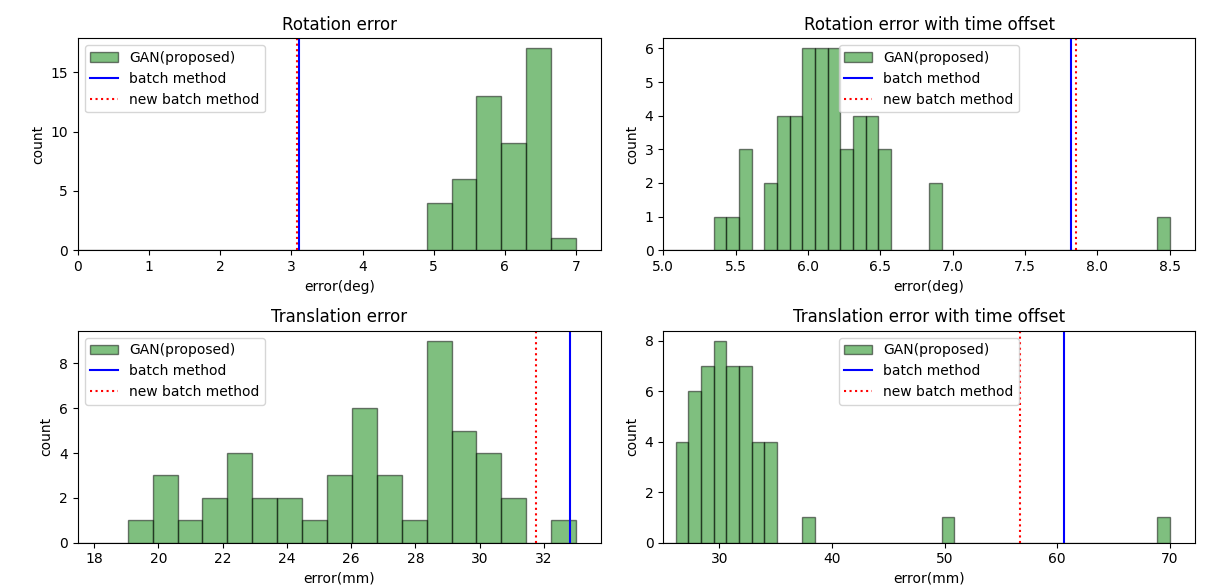}
\caption{Error comparison for Exp.~3 between proposed and existing methods using experimental data without time offsets (left column) and with simulated time offsets (right column).}
\label{fig:exp3}
\end{figure}

The results of the three experiments are shown in the left columns of Fig.~\ref{fig:exp1}, Fig.~\ref{fig:exp2}, and Fig.~\ref{fig:exp3}, respectively. Because of its stochastic nature, we applied our method $50$ times and presented the errors in the histograms. By contrast, the compared methods are deterministic, and their results are presented as vertical solid and dashed lines in the figures. In Exp.~2 and Exp.~3, we observed comparable performances among all methods in rotation errors, while the translation errors of our method at the histogram peaks were significantly smaller than the translation errors of the other methods. In Exp.~1, our method outperformed the others in rotation errors, while the translation errors were comparable across all methods.

Considering that the translation estimation of our method was less accurate than the other methods in the simulation outcomes, the superiority of our method was more pronounced in the hardware experiment than in the simulations. One possible explanation for this is that, in the simulations, datasets were generated using Gaussian distributions, which could be an ideal situation for the analytical methods that rely solely on data mean and covariance. On the other hand, in the hardware experiment, the data distributions were more nonlinear, making their higher-order moments more significant. The analytical methods may not perform optimally in this situation.

In the experiments, our method took $316$ seconds on average for each calibration, which was longer than the average running time in the simulations due to larger data sets and nonlinearity of data distribution. Again, the same Apple silicon M2 was used with CPU computation using PyTorch.

\begin{table}
\caption{Mean and standard deviation of rotation and translation errors in hardware experiments}
\label{tab:exp}
\begin{center}
\scalebox{0.87}{
\begin{tabular}{ l|l|c|c|c|c}
\cline{3-6}
\multicolumn{2}{l}{} & \multicolumn{2}{c|}{\bf Without Time Offset} & \multicolumn{2}{c}{\bf With Time Offset} \\
\cline{3-6}
\multicolumn{2}{l}{} & {\bf Rotation} & {\bf Translation} & {\bf Rotation} & {\bf Translation} \\
\multicolumn{2}{l}{} & {\bf Error ($^\circ$)} & {\bf Error (mm)} & {\bf Error ($^\circ$)} & {\bf Error (mm)} \\
\hline
\hline
\multirow{4}{*}{\bf Exp. 1} & \multirow{2}{*}{\bf GAN (ours)} & {$1.03$} & {$38.06$} & {$0.97$} & {$53.82$} \\
& & {$(\pm 9.13)$} & {$(\pm 33.81)$} & {$(\pm 25.81)$} & {$(\pm 135.85)$} \\
\cline{2-6}
& {\bf Batch} & {$1.99$ } & {$39.83$} & {$6.80$ } & {$83.81$} \\
\cline{2-6}
& {\bf New Batch} & {$1.99$} & {$37.97$} & {$6.80$} & {$79.73$} \\
\hline
\hline
\multirow{4}{*}{\bf Exp. 2} & \multirow{2}{*}{\bf GAN (ours)} & {$1.68$} & {$41.02$} & {$1.87$} & {$53.13$} \\
& & {$(\pm 9.96)$} & {$(\pm 57.13)$} & {$(\pm 17.17)$} & {$(\pm 128.78)$} \\
\cline{2-6}
& {\bf Batch} & {$1.31$ } & {$53.13$} & {$10.13$} & {$176.95$} \\
\cline{2-6}
& {\bf New Batch}& {$1.30$ } & {$52.07$} & {$10.13$} & {$176.09$} \\
\hline
\multirow{4}{*}{\bf Exp. 3} & \multirow{2}{*}{\bf GAN (ours)} & {$5.98$} & {$26.00$} & {$6.00$} & {$30.38$} \\
& & {$(\pm 2.06)$} & {$(\pm 25.95)$} & {$(\pm 4.74)$} & {$(\pm 34.51)$} \\
\cline{2-6}
& {\bf Batch} & {$3.10$ } & {$32.81$} & {$7.81$ } & {$60.49$} \\
\cline{2-6}
& {\bf New Batch} & {$3.07$} & {$31.74$} & {$7.84$} & {$56.72$} \\
\hline
\end{tabular}
}
\end{center}
\end{table}


To demonstrate a practical scenario in which time offsets exist in the start and end times between the two measurement streams, we tested our method and existing methods when the camera measurement stream was assumed to start and end earlier than the robot transformation stream by $5$\% of the total stream time ($=$ approximately $9$ sec). This situation was simulated by discarding the first $5$\% and last $5$\% of the robot and camera measurements, respectively. The results are shown in the right columns of Fig.~\ref{fig:exp1}, Fig.~\ref{fig:exp2}, and Fig.~\ref{fig:exp3}. As expected, the other methods exhibited poorer performance in the presence of time offsets, amplifying the performance gap in comparison to our method. On the other hand, our method resulted in slightly degraded solutions in all three experiments, while it showed better retention of solution accuracy against the time offsets than the other methods. Once again, we attribute this result to the fact that our method performs calibration by aligning the entire data distributions.

The means and stds of errors in all cases can also be found in Table~\ref{tab:exp}. Note that the compared methods are deterministic and have no stds.

\section{General Model Parameter Estimation without Data Correspondence}
\label{sec:general_param}

This section presents the generalized procedure for extending the proposed method to model parameter estimation, along with the condition under which the extension is feasible. Finally, a simple example will be provided to demonstrate polynomial fitting without data correspondence.

Let $\{ a_i \}$ and $\{ b_j \}$ denote the datasets without correspondences. Assuming that they are not necessarily the inputs and outputs of the underlying model, the model can be written in a general form as
\begin{equation}
f(a, b, x) = 0 \label{eqn:f}
\end{equation}
where $x$ is the vector of model parameters.

Note that our approach involves a fixed distribution  and a moving distribution that varies with $X$ (i.e., $\{ B_j \}$ and $\{ X^{-1} A_i X \}$, respectively).
To extend our approach to (\ref{eqn:f}), we need to find an equation that decouples $a$ and $b$ on each side, one of which should be fixed with respect to $x$. Consequently, the condition for our approach to be extendable  is that the model equation can be expressed as
\begin{equation}
g_b(b) = g_a(a, x). \label{eqn:g}
\end{equation}
In other words, $b$ should be algebraically isolated from $a$ and $x$ (or $a$ and $b$ can be swapped). Finally, the GAN-based parameter estimation can be illustrated by replacing $A, B$ and $G$ in Fig.~\ref{fig:method} with $a, g_b(b)$ and $g_a$, respectively.

\subsection{Example: Polynomial Fitting}

\begin{figure}[t!]
  \center
    \includegraphics[width=0.80\columnwidth]{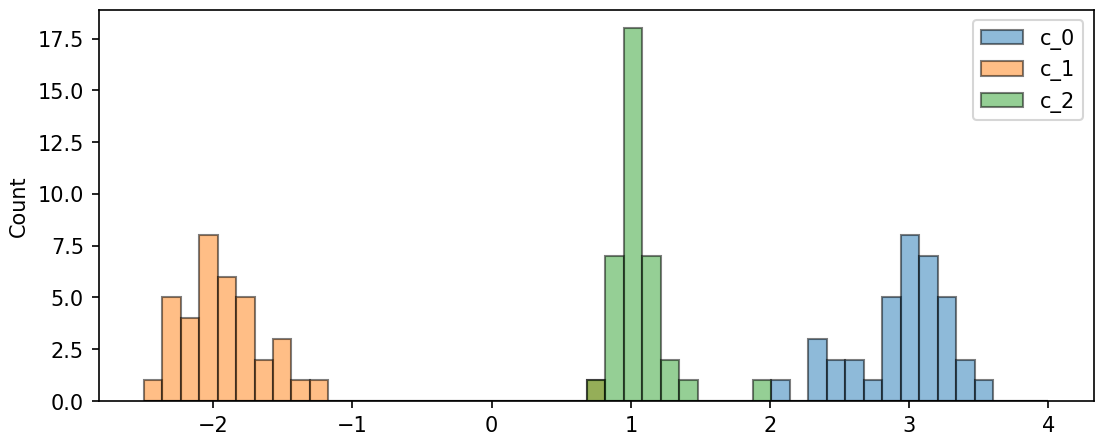} 
\caption{Polynomial parameter estimation result. The ground-truth values are $c_0 = 3, c_1 = -2$, and $c_2 = 1$.}
\label{fig:polynomial}
\end{figure}

As an example, consider the following polynomial function:
\begin{equation}
b = c_0 + c_1 a + c_2 a^2
\label{eqn:polynomial}
\end{equation}
where $x = (c_0, c_1, c_2)$ denotes the unknown parameters. The equation $g_b(b) = g_a(a,x)$ is available, with $g_a(a,c)$ and $g_b(b)$ given by
\begin{eqnarray}
g_a(a,x) &=& c_0 + c_1 a + c_2 a^2, \\
g_b(b) &=& b.
\end{eqnarray}
These functions can be used for the proposed extension. As a test, we generated $400$ random $a$'s and $600$ random $b$'s for a uniform distribution of $a$ in $a \in [-4, 3]$ given the true parameters $x = (3,-2,1)$. The estimated $c_1, c_2$, and $c_3$ values from $100$ independent executions are presented as histograms in Fig.~\ref{fig:polynomial}. As observed, the estimations were distributed around their true values, confirming the feasibility of extending our method to general parameter estimations.

\section{Conclusions}
\label{sec:conclusions}

In this study, we rediscovered the GAN training framework as a solver for hand-eye calibration without data correspondence. The proposed method has some desirable properties inherited from the established benefits of GANs, including the ease of handling nonlinear manifolds, alignment between entire distributions, and extendability to general parameter estimations. Most significantly, this study opens a novel utility of GANs that has not been explored conventionally.

In terms of performance, it outperformed existing analytical methods in our hardware experiment, particularly when the datasets contained time offsets. Meanwhile, solution convergence can be degraded by the stochasticity of GAN training. For this reason, our future studies will focus on improving convergence and extending the method to various manifold problems, including the $AX=YB$ and $AXB=YCZ$ problems.


	
	%

	


	\bibliographystyle{IEEEtran}
	\bibliography{references}

	
	
\end{document}